\documentclass{article}

\usepackage{microtype}
\usepackage{graphicx}
\usepackage{subcaption}
\usepackage{booktabs} 
\usepackage[table]{xcolor}

\usepackage[most]{tcolorbox}

\usepackage{hyperref}
\usepackage{pifont}
\usepackage{xspace}

\definecolor{lblu_tab}{RGB}{225, 235, 246}
\definecolor{orange_vitad}{RGB}{222, 131, 68}
\definecolor{blue_vitad}{RGB}{106, 153, 208}
\definecolor{trajectory_green}{RGB}{126, 171, 85}
\definecolor{trajectory_yellow}{RGB}{245, 194, 66}
\definecolor{tab_others}{RGB}{235, 235, 235}
\definecolor{tab_ours}{RGB}{225, 235, 246}
\definecolor{lightblue}{rgb}{0.93, 0.96, 1.0}

\newcounter{prompt}
\renewcommand{\theprompt}{\arabic{prompt}}
\newcommand{\prompt}[3]{
\refstepcounter{prompt}
\begin{tcolorbox}[
    colback=lightblue!35, 
    colframe=white!45!black, 
    title={Prompt.~\theprompt:~#1},
    breakable,
]
#2
\label{#3}
\end{tcolorbox}
}

\usepackage[preprint]{icml2026}

\usepackage{amsmath}
\usepackage{amssymb}
\usepackage{mathtools}
\usepackage{amsthm}
\usepackage{makecell}

\usepackage[capitalize,noabbrev]{cleveref}
\newcommand{\modelname}{Kiwi-Edit}
\newcommand{\datasetname}{RefVIE}
\theoremstyle{plain}

\theoremstyle{definition}

\theoremstyle{remark}

\definecolor{whit_tab}{RGB}{255, 255, 255}
\definecolor{gray_tab}{RGB}{246, 246, 246}
\definecolor{oran_tab}{RGB}{252, 242, 237}
\definecolor{blue_tab}{RGB}{227, 240, 251}

\newcommand{\weburl}{Project Page: \url{https://showlab.github.io/Kiwi-Edit}\xspace}

\usepackage[textsize=tiny]{todonotes}
\usepackage{bbding}
\icmltitlerunning{\textcolor{scolor}{K}\textcolor{hcolor}{i}\textcolor{ocolor}{w}\textcolor{wcolor}{i}-Edit: Versatile Video Editing via Instruction and Reference Guidance}
\definecolor{scolor}{RGB}{111,168,220}
\definecolor{hcolor}{RGB}{111,176,81}
\definecolor{ocolor}{RGB}{224,103,102}
\definecolor{wcolor}{RGB}{246,178,107}

\begin{document}

\twocolumn[
  \icmltitle{\includegraphics[height=2ex]{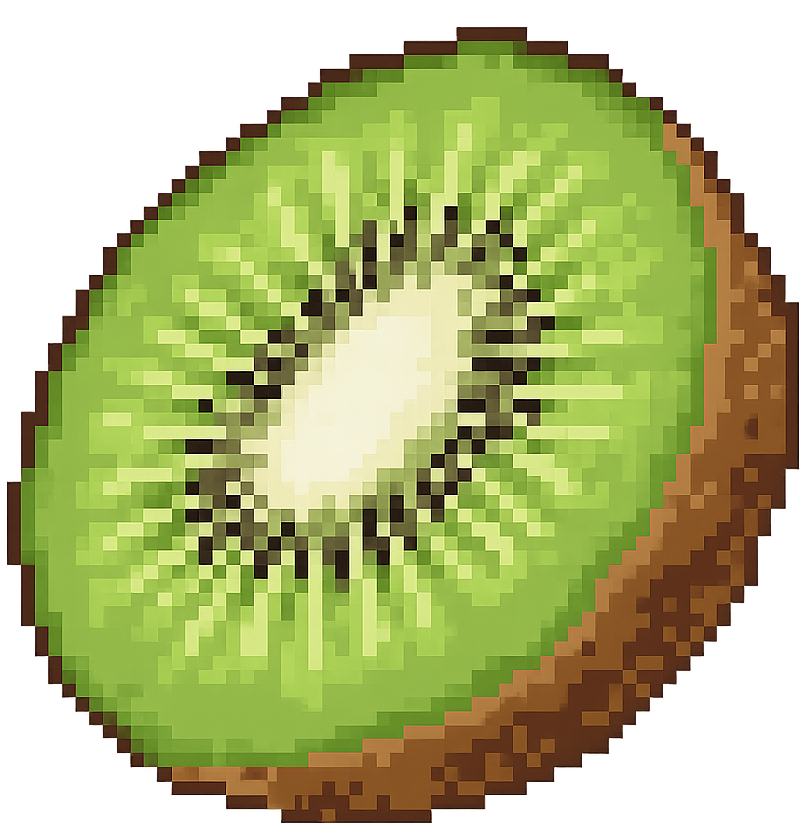}\textcolor{scolor}{K}\textcolor{hcolor}{i}\textcolor{ocolor}{w}\textcolor{wcolor}{i}-Edit: Versatile Video Editing via Instruction and Reference Guidance}



  \icmlsetsymbol{co}{\Envelope}
  \begin{icmlauthorlist}
    \icmlauthor{Yiqi Lin}{}
    \icmlauthor{Guoqiang Liang}{}
    \icmlauthor{Ziyun Zeng}{}
    \icmlauthor{Zechen Bai}{}
    \icmlauthor{Yanzhe Chen}{}
    \icmlauthor{Mike Zheng Shou}{co}
  \end{icmlauthorlist}

    \icmlcorrespondingauthor{Mike Zheng Shou}{mike.zheng.shou @gmail.com}
    \begin{center}
    Show Lab, National University of Singapore
    
    \weburl
    \end{center}
  \icmlkeywords{Instruction Video Editing}

  \vskip 0.3in

  \begin{center}
  \vspace{-12pt}\centerline{\includegraphics[width=\textwidth]{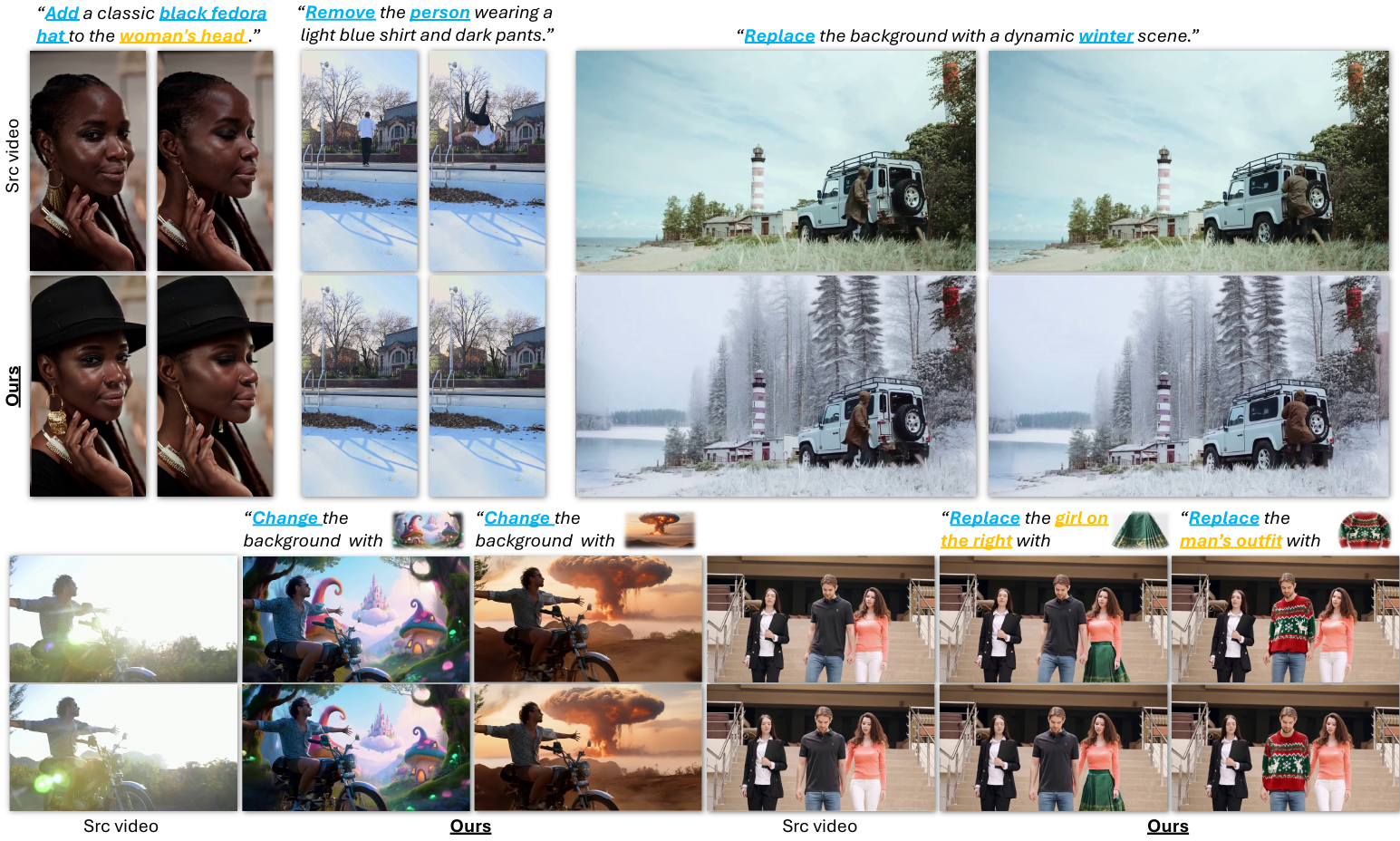}}
    \captionof{figure}{This teaser illustrates a selection of video editing tasks, including both instruction-only and instruction-reference scenarios, highlighting the superior editing capabilities of \datasetname.}
    \label{fig:teaser}
  \end{center}
]



\printAffiliationsAndNotice{}  

\begin{abstract}
Instruction-based video editing has witnessed rapid progress, yet current methods often struggle with precise visual control, as natural language is inherently limited in describing complex visual nuances. Although reference-guided editing offers a robust solution, its potential is currently bottlenecked by the scarcity of high-quality paired training data. To bridge this gap, we introduce a scalable data generation pipeline that transforms existing video editing pairs into high-fidelity training quadruplets, leveraging image generative models to create synthesized reference scaffolds. Using this pipeline, we construct {RefVIE}, a large-scale dataset tailored for instruction-reference-following tasks, and establish {RefVIE-Bench} for comprehensive evaluation. Furthermore, we propose a unified editing architecture, \modelname, that synergizes {learnable queries} and {latent visual features} for reference semantic guidance. Our model achieves significant gains in instruction following and reference fidelity via a progressive multi-stage training curriculum. Extensive experiments demonstrate that our data and architecture establish a new state-of-the-art in controllable video editing. \textit{All datasets, models, and code is released at \url{https://github.com/showlab/Kiwi-Edit}.}
\end{abstract}

%
\section{Introduction}
\label{sec:intro}

The customization of video content across social media, entertainment, and advertising has fueled an unprecedented demand for accessible video editing tools. 
Recent advances in instruction-based video editing~\cite{cheng2023consistent,ditto,zi2025se,wu2025insvie,openve} have demonstrated remarkable progress, enabling users to modify video content through natural language commands. 
These methods leverage powerful video diffusion models~\cite{wan2025wan,kong2024hunyuanvideo} to execute diverse editing operations, ranging from local object modifications to global style transfers, while preserving the content and temporal coherence across frames.

Despite this progress, a critical limitation persists: the reliance on text-only instructions. 
Natural language is inherently ambiguous when describing precise visual details such as specific textures, exact object identities, or nuanced stylistic characteristics. 
Users frequently desire to convey editing intent through visual examples, such as ``replace the car with \textit{this} sports car'' or ``apply the style of \textit{this} painting,'' yet text-only models fundamentally struggle to accomplish such tasks. 
Reference-guided video editing, which conditions generation on both textual instructions and visual references, offers a natural solution to this challenge.

However, the development of reference-guided video editing is severely constrained by data scarcity. 
Training such models requires high-quality quadruplets comprising source videos, editing instructions, reference images, and target videos, a format that existing datasets do not provide at scale. 
As summarized in \cref{tab:dataset_comparison}, while several large-scale instruction-based video editing datasets exist~\cite{wu2025insvie,zi2025se,ditto,Reco,openve}, \emph{none offer reference image}.
The few works that explore reference-guided editing~\cite{mou2025instructx,team2025kling} rely on proprietary data that remain inaccessible to the wider research community.
This bottleneck fundamentally impedes the field.

To address this challenge, we curate {\datasetname}, a large-scale dataset for instruction-reference guided video editing. 
Our key insight is that powerful pre-trained image generation models can serve as high-fidelity reference synthesizers, enabling scalable data construction without expensive manual annotation. 
Starting from existing instruction-based video editing datasets~\cite{ditto,Reco,openve} that provide source-target video pairs, we design an automated pipeline to synthesize the missing reference images.
Specifically, we leverage vision-language models to ground editing regions in video frames, followed by state-of-the-art image editors to generate reference images that capture the visual essence of the intended edit. 
Through rigorous quality filtering and de-duplication, we construct a dataset of \textbf{477K} high-quality quadruplets from an initial pool of 3.7M samples. To the best of our knowledge, \textbf{\datasetname~is the first large-scale, open-source resource for instruction-reference guided video editing.}

Building upon this data foundation, we develop \textbf{\modelname}, a unified video editing framework that effectively integrates multimodal conditions. 
Our architecture couples a frozen Multimodal Large Language Model (MLLM) with a Diffusion Transformer (DiT), where the MLLM processes interleaved sequences of source video frames, textual instructions, and reference images. 
We employ a dual-connector mechanism: a \textit{Query Connector} that projects learnable query tokens to distill editing intent, and a\textit{ Latent Connector} that extracts visual features from reference images. 
These connectors produce unified context tokens that guide the DiT via cross-attention.
To preserve source video structure while enabling flexible reference-guided editing, we introduce a hybrid latent injection strategy: source video features are added element-wise with a timestep-dependent scalar for structural preservation, while reference image features are concatenated to the input sequence for fine-grained texture transfer.
A three-stage curriculum training strategy ensures stable convergence: MLLM-DiT alignment, instructional tuning, and reference-guided fine-tuning. 
Furthermore, to enable rigorous evaluation, we establish {RefVIE-Bench}, a benchmark of 100 manually verified samples designed to assess reference adherence, instruction compliance, and temporal consistency.

In summary, our contributions are threefold:
First, we curate {\datasetname}, a large-scale dataset of 477K high-quality quadruplets for instruction-reference guided video editing, covering local editing and background replacement. 
Second, we introduce {RefVIE-Bench}, a comprehensive benchmark specifically designed to evaluate reference similarity, instruction accuracy, and temporal consistency.
Lastly, we present a unified video editing model that achieves state-of-the-art performance on both instruction-only and reference-guided tasks through a novel MLLM-DiT architecture design and multi-stage training curriculum.

\begin{table}
    \centering
    \caption{A summary of existing large scale instruction and reference guided video editing datasets.}
    \resizebox{1\linewidth}{!}{
\begin{tabular}{cccccc}
\hline
Dataset & Open-Source & Instr. Edit & Ref. Image & Num. \\ \hline
InsViE~\cite{wu2025insvie} & \textcolor{green}{\ding{51}} & \textcolor{green}{\ding{51}} & \textcolor{red}{\ding{55}} & 1M  \\
Señorita~\cite{zi2025se} & \textcolor{green}{\ding{51}} & \textcolor{green}{\ding{51}} & \textcolor{red}{\ding{55}} & 2M  \\
Ditto~\cite{ditto} & \textcolor{green}{\ding{51}} & \textcolor{green}{\ding{51}} & \textcolor{red}{\ding{55}} & 1M  \\
ReCo~\cite{Reco} & \textcolor{green}{\ding{51}}   & \textcolor{green}{\ding{51}} & \textcolor{red}{\ding{55}} & 500K          \\
OpenVE~\cite{openve} & \textcolor{green}{\ding{51}} & \textcolor{green}{\ding{51}} & \textcolor{red}{\ding{55}}      & 3M  \\
\bottomrule
InsturctX~\cite{mou2025instructx} & \textcolor{red}{\ding{55}} & \textcolor{green}{\ding{51}} & \textcolor{green}{\ding{51}}  & 236K \\
Kling-Omni~\cite{team2025kling} & \textcolor{red}{\ding{55}} & \textcolor{green}{\ding{51}} & \textcolor{green}{\ding{51}}  & --- \\
RefVIE (Ours)  &\textcolor{green}{\ding{51}} & \textcolor{green}{\ding{51}}  & \textcolor{green}{\ding{51}}   & 477K          \\ \hline
\end{tabular}
}
\vspace{-10pt}
\label{tab:dataset_comparison}
\end{table}

\vspace{-10pt}

\section{Related Work}

\begin{figure*}
    \centering
    \includegraphics[width=0.9\linewidth]{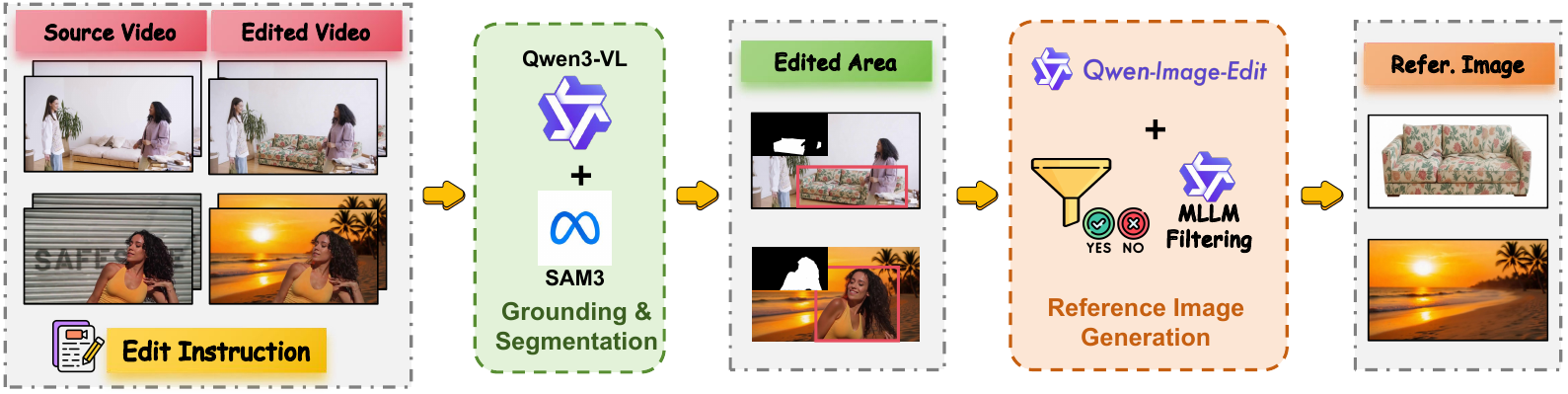}
\vspace{-5pt}
\caption{Workflow of the reference image synthesis pipeline. We first ground the editing region in the target video frame using specialized grounding and segmentation models. Subsequently, we leverage a specialized image editing model to synthesize a high-quality reference image that maintains identity consistency with the instruction.}
    \label{fig:workflow}
\vspace{-8pt}
\end{figure*}

\subsection{Instruction-based Video Editing}
Given the challenges in training robust text-to-video models from scratch, early prevalent approaches~\cite{wu2023tune,cong2023flatten,geyer2023tokenflow,kara2024rave,ku2024anyv2v,qi2023fatezero} leverage the rich generative priors of pre-trained text-to-image (T2I) models. These methods typically employ fine-tuning or inversion techniques~\cite{song2020denoising} to achieve instruction-guided editing. 
However, T2I-based methods often suffer from limited temporal consistency and inversion artifacts, particularly under complex motion or occlusions. Consequently, with the advent of open-source video diffusion models~\cite{hong2022cogvideo,kong2024hunyuanvideo,wan2025wan}, recent research has shifted towards utilizing these native video backbones to ensure better motion fidelity.
To support instruction-based training, InsV2V~\cite{cheng2023consistent} pioneers the field by utilizing InstructPix2Pix~\cite{brooks2023instructpix2pix} to synthesize paired training data. Addressing the fundamental challenge of data scarcity, subsequent efforts~\cite{ditto,zi2025se,wu2025insvie} focus on constructing large-scale synthesized datasets. For instance, Senorita-2M~\cite{zi2025se} collects editing pairs via a mixture-of-experts pipeline, while Ditto~\cite{ditto} leverages edited key-frames and depth maps to generate edited videos. 
More recently, to enhance instruction following and semantic understanding, Omni-Video~\cite{tan2025omni} and OpenVE-Edit~\cite{openve} integrate Vision-Language Models (VLMs) into the editing framework.

\subsection{Reference-Guided Video Editing and Dataset}
Relying solely on natural language prompts often fails to capture the nuances of visual imagination. Text is inherently limited in describing precise spatial relationships, specific visual references, and temporal dynamics, creating a gap between user intent and model output. 
To address this, recent works~\cite{mou2025instructx,wei2025univideo,team2025kling} have introduced reference images alongside textual instructions to enable more precise video editing. Specifically, methods like InstructX~\cite{mou2025instructx} and Kling-Omni~\cite{team2025kling} feed multi-modal inputs into MLLMs to extract unified representations for the generation module. 
However, despite their impressive performance, these approaches rely heavily on proprietary in-house models for data generation and require extensive manual verification to curate high-quality reference data. 
A summary of represented video editing datasets is presented in \cref{tab:dataset_comparison}.
In this work, we aim to democratize this capability by introducing \textbf{\datasetname}, the first large-scale open-source dataset tailored for instruction-reference guided video editing. By providing a rigorously filtered and curated dataset, we offer a critical resource to the research community, enabling the development of comprehensive editing models that effectively transcend the limitations of text-only control.

\vspace{-10pt}

\section{\datasetname~Dataset and Benchmark}

\subsection{Scalable Data Generation Pipeline}
\label{sec:data_pipeline}

\begin{figure}
    \centering
    \includegraphics[width=\linewidth]{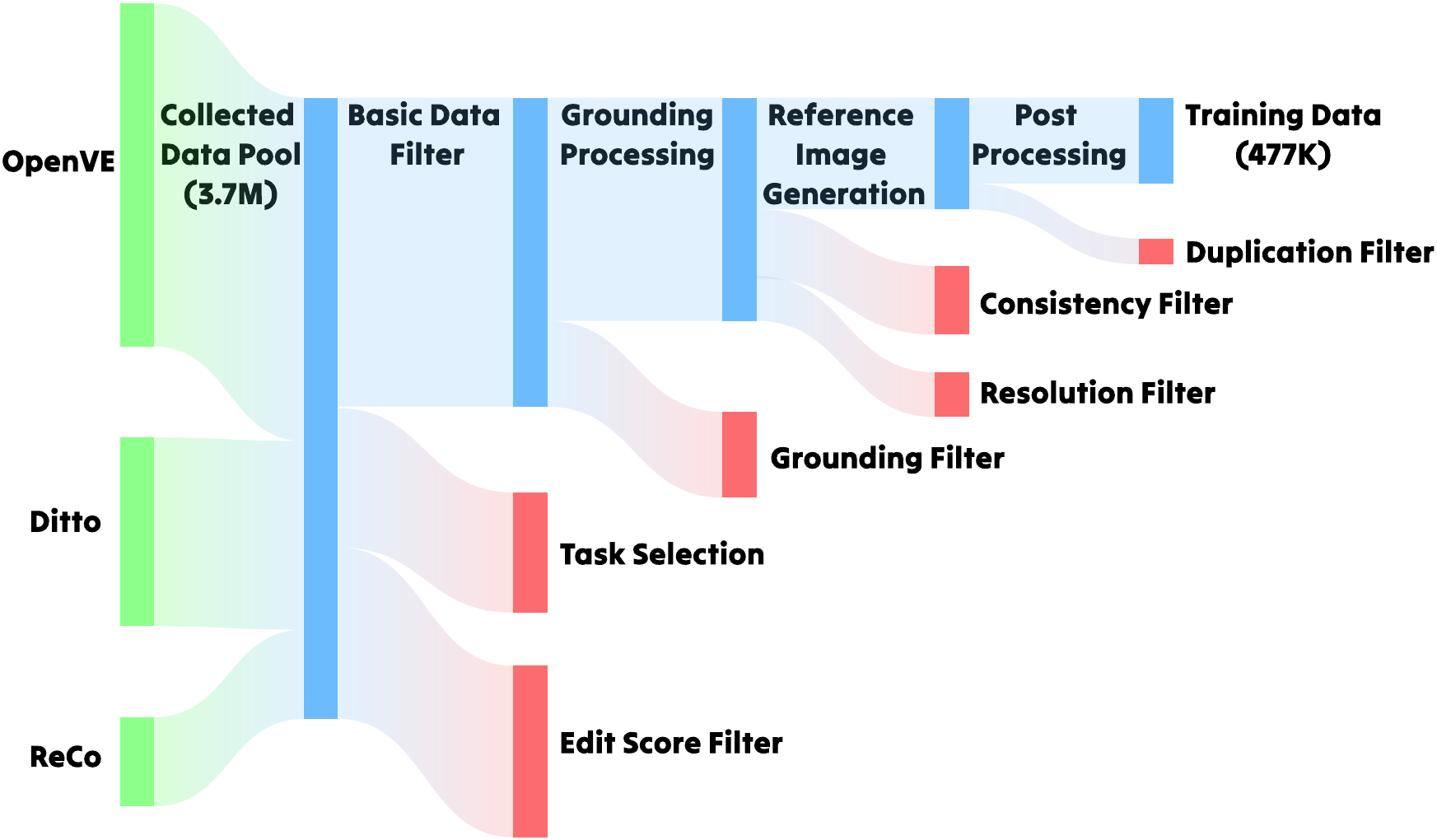}
    \vspace{-5pt}
    \caption{Pipeline of \datasetname~curation. We process 3.7M raw samples through four stages: source aggregation and filtering, grounding and segmentation, reference image synthesis, and quality control, yielding 477K high-quality quadruplets.}
    \vspace{-15pt}
    \label{fig:datapipe}
\end{figure}

A primary bottleneck in advancing reference-guided video editing is the scarcity of high-quality training quadruplets: $(V_{src}, T_{inst}, I_{ref}, V_{tgt})$. Manual curation of such 4-tuple data is prohibitively expensive. To address this, we propose a scalable, automated pipeline to synthesize reference images from existing instruction-based video editing pairs, effectively augmenting standard triplets $(V_{src}, T_{inst}, V_{tgt})$ into the required quadruplets. As illustrated in \cref{fig:datapipe}, our pipeline processes a massive pool of 3.7M raw samples from publicly available datasets through four distinct stages.

\noindent\textbf{Stage 1. Source Aggregation and Filtering.} 
We initialize our data pool by aggregating three open-source instructional video editing datasets: Ditto-1M~\cite{ditto}, ReCo~\cite{Reco}, and OpenVE-3M~\cite{openve}. To ensure high training quality, we filter these samples using EditScore~\cite{luo2025editscore}. Based on a human-pivoted pilot study, we discard samples with an EditScore lower than 6 for text-guided instruction tuning. For reference-guided generation specifically, we apply a stricter threshold (EditScore $> 8$) and explicitly select tasks categorized as \textit{Local Modification} or \textit{Background Replacement}, as these benefit most from visual referencing.

\noindent\textbf{Stage 2. Grounding and Segmentation.} 
Precise spatial localization is critical for generating consistent references. We employ Qwen3-VL-32B~\cite{qwen3vl} to interpret the editing instruction and ground the region of interest in the first frame of the \textit{target} video, as the target video contains the desired editing result from which we extract the reference.
For \textit{Background Change}, the model grounds the foreground object so that it can be removed in the next stage, leaving only the new background as the reference.
For \textit{Local Editing}, the model grounds the edited object so that it can be extracted as the reference.
These coarse bounding box coordinates are then refined by SAM3~\cite{sam3} to produce pixel-perfect segmentation masks. Pairs that fail the grounding or segmentation checks are discarded.

\noindent\textbf{Stage 3. Reference Image Synthesis.} 
Leveraging the segmented regions, we synthesize reference images using Qwen-Image-Edit-2511~\cite{qwenimage}, as depicted in \cref{fig:workflow}.
For background tasks, we extract and remove the foreground object, then inpaint the region to produce a clean background image that serves as the reference.
For local edits, we extract the target object and place it on a clean background with minimal surrounding space, creating a tightly cropped reference that highlights the edited object's appearance.
To ensure data robustness, we also filter out generated images with extreme aspect ratios or resolution.

\noindent\textbf{Stage 4. Quality Control and Post-Processing.} 
In the final stage, we enforce semantic alignment by using an MLLM to verify that the synthesized reference image is consistent with the edited content in the target video, filtering out low-fidelity generations. Additionally, to prevent data leakage and redundancy, we extract CLIP~\cite{radford2021learning} features from the reference images and perform global de-duplication. 
This rigorous pipeline distills the initial 3.7M pool down to a high-quality subset of 477K instruction-reference-video quadruplets. The detailed breakdown across filtering stages is visualized in \cref{fig:datapipe}.

\subsection{Dataset Statistics}
\label{sec:data_stats}

\begin{figure}
    \centering
    \includegraphics[width=\linewidth]{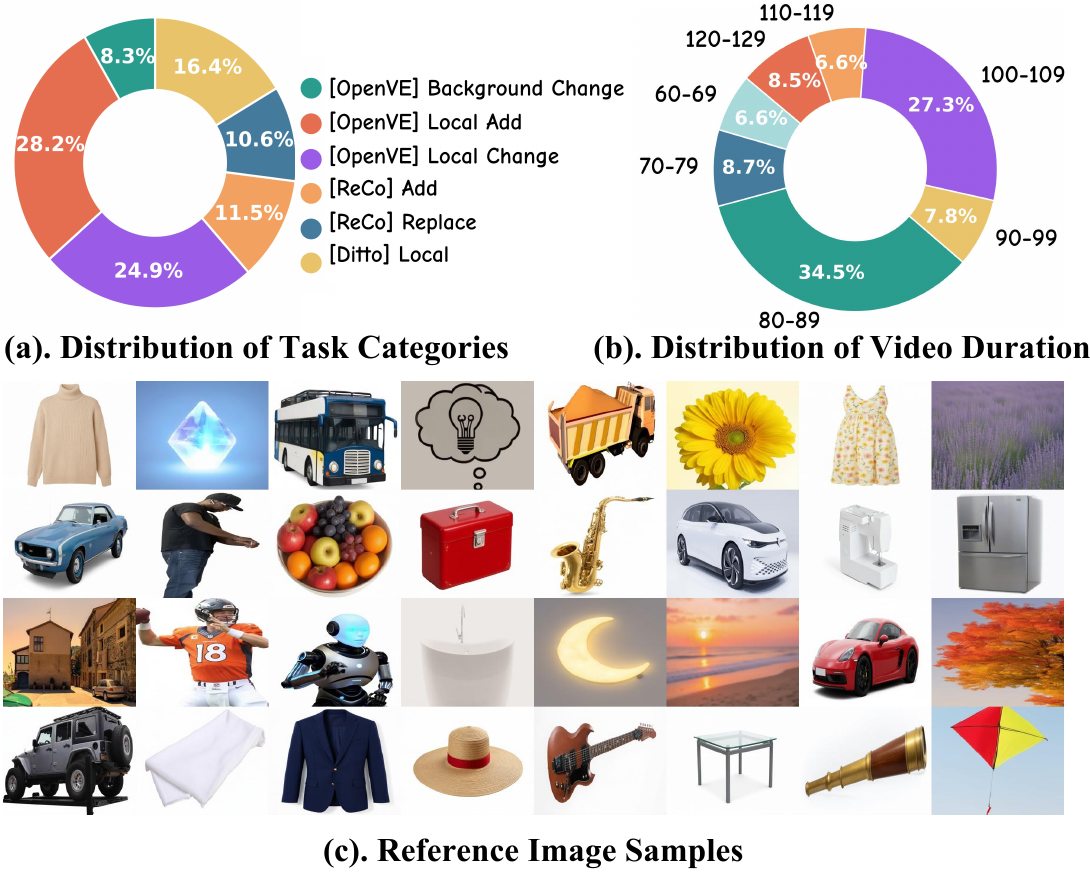}
    \vspace{-10pt}
    \caption{\datasetname~statistics and sample visualization. (a) Distribution of editing task types. (b) Distribution of video durations. (c) Example reference images for different editing categories.}
    \label{fig:stat}
    \vspace{-15pt}
\end{figure}

Our resulting dataset, \textbf{\datasetname}, is the largest open-source collection for reference-guided video editing, bridging the gap between academic resources and commercial capabilities.
\cref{fig:stat} summarizes the dataset statistics. 
As shown in \cref{fig:stat}(a), the task distribution is well-balanced across local object addition, replacement, and background changes, ensuring that trained models generalize across different editing scenarios rather than overfitting to a single task type.
\cref{fig:stat}(b) illustrates the video duration distribution. Most clips contain 80 to 110 frames, providing sufficient temporal context for models to learn long-range motion consistency and handle complex object movements.
\cref{fig:stat}(c) presents example reference images, demonstrating the diversity of visual content including objects, textures, and backgrounds.
More visualization cases are provided in the supplementary.

\begin{figure*}
    \centering
    \includegraphics[width=0.92\linewidth]{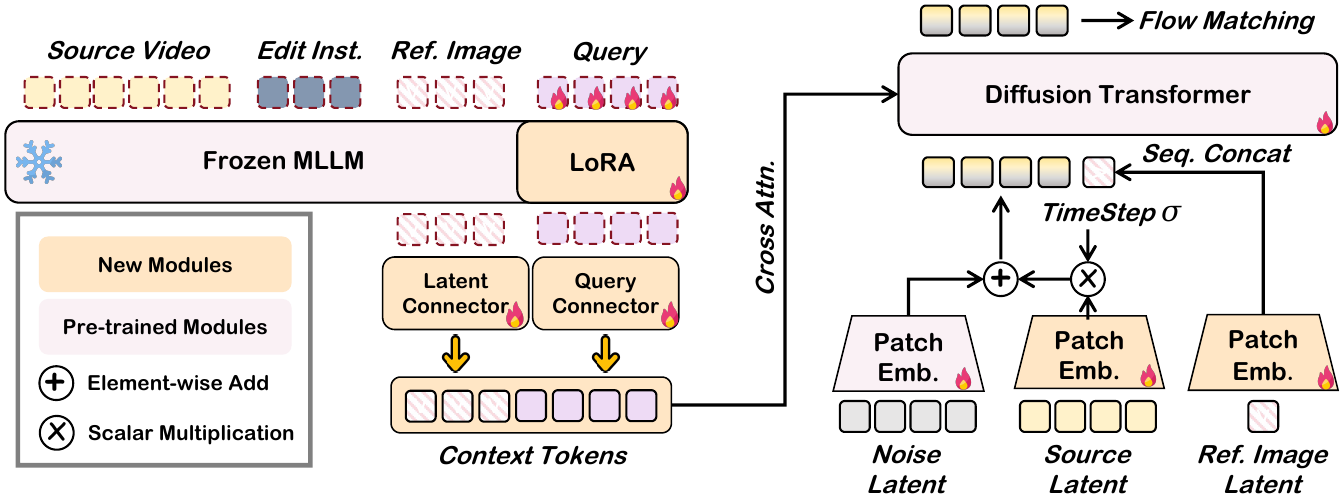}
\caption{Overview of our unified editing framework. We integrate a frozen MLLM (Qwen2.5-VL-3B) to encode multimodal instructions, injecting semantic conditions into the pre-trained Diffusion Transformer (Wan2.2-TI2V-5B) via dual learnable projectors for query and reference latents. To preserve consistency of source video, we employ a hybrid injection strategy within the DiT: source video features are added element-wise, while reference image features are concatenated to the input sequence.}
    \label{fig:model}
\vspace{-5pt}
\end{figure*}

\subsection{Benchmark and Evaluation}
\label{sec:benchmark}

\textbf{RefVIE-Bench.} Existing benchmarks predominantly focus on text-video alignment, neglecting the critical dimension of visual reference adherence. To rigorously evaluate this capability, we establish {RefVIE-Bench}, comprising 110 manually verified triplets $(V_{src}, I_{ref}, T_{ins})$.
Unlike our scalable synthetic training data, these benchmark samples undergo a rigorous three-stage manual verification process to ensure high quality and diversity.
The benchmark evaluates two specific capabilities: \textit{Subject Reference} (70 samples) and \textit{Background Replacement} (40 samples). The larger allocation to object modification reflects its broader scope, encompassing diverse object categories such as vehicles, animals, clothing, and furniture, each with distinct visual attributes. In contrast, Background replacement represents a more unified task category focused on environment changes while preserving foreground dynamics.

\textbf{Evaluation Metrics.} Traditional metrics like CLIP score only capture high-level semantic similarity, while FID measures distribution-level statistics rather than per-sample quality. Neither can assess whether specific textures are preserved or whether editing instructions are correctly followed. To address this, we employ a state-of-the-art MLLM (Gemini3~\cite{comanici2025gemini}) as an automated judge, which evaluates each result on a scale of 1 to 5 across three dimensions. 
For subject reference, we evaluate \textit{Identity Consistency}, \textit{Temporal Fidelity}, and \textit{Physical Integration} (e.g., tracking, shadows), while for background replacement, we assess \textit{Reference Fidelity}, \textit{Matting Quality}, and \textit{Visual Harmony} (e.g., perspective, lighting). To ensure logical robustness, we enforce a hierarchical constraint where temporal and physical scores are capped by the primary identity score, preventing the model from assigning high ratings to edits that are temporally stable but semantically incorrect.

\section{Methodology}

\subsection{Architecture Design}
\label{sec:arch}

As illustrated in \cref{fig:model}, our framework consists of two main components: a Multimodal Large Language Model (MLLM) for semantic understanding and a Diffusion Transformer (DiT) for video generation. The MLLM encodes multimodal inputs (source video, instruction, and optional reference image) into conditioning signals, which guide the DiT to generate the edited video. 

\noindent\textbf{Semantic Conditioning via MLLM.} 
We utilize Qwen2.5-VL-3B~\cite{bai2025qwen2} as the MLLM backbone. The base model weights are frozen, and we inject lightweight Low-Rank Adaptation~\cite{hu2022lora} (LoRA) modules to adapt it to the video editing domain without compromising pre-trained knowledge.
The MLLM processes an interleaved sequence comprising the source video frames, textual editing instructions, and optional reference images. From its output, we extract conditioning features through two specialized pathways:
\textit{Instructional Queries:} We utilize a set of learnable query tokens (256 for image tasks, 512 for video editing, 768 for reference task) to distill the editing intent (e.g., ``turn the sky red''). These are projected via a \textit{Query Connector} (MLP) to align with the DiT's dimension.
\textit{Reference Latents:} For tasks requiring specific visual guidance, we extract visual tokens corresponding to the reference image. These are projected via a separate \textit{Latent Connector}.
The outputs of these connectors are concatenated to form a unified sequence of \textit{Context Tokens}, which serve as the key/value pairs for the DiT's cross-attention layers, guiding the semantic content of the generation.

\begin{table*}
\centering
\caption{OpenVE-Bench results evaluated on Gemini-2.5-Pro. Orange rows denote closed-source models and are excluded when selecting the best and second-best scores; gray rows denote open-source baselines and blue rows denote our variants. Bold and underlined scores indicate the best and second-best results among non-closed-source methods, respectively.}
\label{tab:openve}
\resizebox{\textwidth}{!}{%
\begin{tabular}{l|cc|c|ccccc}
\toprule
\textbf{Method} & \makecell[c]{\textbf{\#Params.}\\\textbf{(DiT)}} & \makecell[c]{\textbf{\#Reso.}\\} & \makecell[c]{\textbf{Overall}$\uparrow$\\} & \makecell[c]{\textbf{Global}\\\textbf{Style} $\uparrow$} & \makecell[c]{\textbf{Background}\\\textbf{Change} $\uparrow$} & \makecell[c]{\textbf{Local}\\\textbf{Change} $\uparrow$} & \makecell[c]{\textbf{Local}\\\textbf{Remove} $\uparrow$} & \makecell[c]{\textbf{Local}\\\textbf{Add} $\uparrow$} \\ \midrule
\rowcolor{oran_tab}\textcolor{black!65}{Runway Aleph} & \textcolor{black!65}{-} & \textcolor{black!65}{1280$\times$720} & \textcolor{black!65}{3.49} & \textcolor{black!65}{3.72} & \textcolor{black!65}{2.62} & \textcolor{black!65}{4.18} & \textcolor{black!65}{4.16} & \textcolor{black!65}{2.78} \\
\rowcolor{gray_tab}VACE~\cite{jiang2025vace} & 14B & 1280$\times$720 & 1.57 & 1.49 & 1.55 & 2.07 & 1.46 & 1.26 \\
\rowcolor{gray_tab}OmniVideo~\cite{tan2025omni} & 1.3B & 640$\times$352 & 1.19 & 1.11 & 1.18 & 1.14 & 1.14 & 1.36 \\
\rowcolor{gray_tab}InsViE~\cite{wu2025insvie} & 2B & 720$\times$480 & 1.45 & 2.20 & 1.06 & 1.48 & 1.36 & 1.17 \\
\rowcolor{gray_tab}Lucy-Edit~\cite{lucyedit} & 5B & 1280$\times$704 & 2.22 & 2.27 & 1.57 & 3.20 & 1.75 & 2.30 \\
\rowcolor{gray_tab}ICVE~\cite{liao2025context} & 13B & 384$\times$240 & 2.18 & 2.22 & 1.62 & 2.57 & 2.51 & 1.97 \\
\rowcolor{gray_tab}DITTO~\cite{ditto} & 14B & 832$\times$480 & 2.13 & \textbf{4.01} & 1.68 & 2.03 & 1.53 & 1.41 \\
\rowcolor{gray_tab}OpenVE-Edit~\cite{openve} & 5B & 1280$\times$704 & 2.50 & 3.16 & 2.36 & 2.98 & 1.85 & 2.15 \\
\rowcolor{gray_tab}ReCo~\cite{Reco} & 2.1B & 832$\times$480 & 2.80 & \underline{3.96} & 1.92 & 3.70 & 2.24 & 2.17 \\
\rowcolor{gray_tab}UniVideo~\cite{wei2025univideo} & 14B & 854$\times$480 & \underline{3.02} & 3.67 & 2.51 & 3.79 & \textbf{2.73} & \underline{2.40} \\
\midrule
\rowcolor{blue_tab}
\textbf{Ours} (Stage-2 Instruct-Only) & 5B & $720\times480$ & 2.92 & 3.54 & \underline{2.59} & 3.80 & 2.55 & 2.12\\
\rowcolor{blue_tab}
\textbf{Ours} (Stage-2 Instruct-Only) & 5B & $1280\times704$ & 2.98 & 3.54 & 2.57 & \underline{3.84} & \underline{2.71} & 2.25 \\
\rowcolor{blue_tab}
\textbf{Ours} (Stage-3 Instruct-Reference) & 5B & $1280\times704$ & \textbf{3.11} & 3.72 & \textbf{2.67} & \textbf{3.91} & 2.69 & \textbf{2.55}\\
\bottomrule
\end{tabular}
}
\vspace{-10pt}
\end{table*}

\noindent\textbf{Structural Conditioning via Latent Injection.}
While MLLM context provides semantic guidance, preserving the precise structural layout of the source video requires a more direct signal. We identify that standard cross-attention is insufficient for fine-grained spatial preservation. Therefore, we introduce a hybrid injection strategy.

\textit{Source Video Control (Element-wise Injection):} 
To preserve the spatial-temporal structure, we encode the source frames into latent space using the VAE. These latents are processed by a zero-initialized \texttt{PatchEmbed} layer. As depicted in \cref{fig:model}, rather than concatenating these features (which we found leads to training instability), we add them element-wise to the noisy latent $\mathbf{z}_t$. Crucially, this addition is modulated by a time-dependent factor $\sigma(t)$:
\begin{equation}
\begin{aligned}
    \mathbf{z}'_t ={}& \texttt{PatchEmbed}(\mathbf{z}_t) \\
    &+ \sigma(t) \cdot \texttt{PatchEmbed}_{src}(\mathrm{VAE}(\mathbf{x}_{src})).
\end{aligned}
\label{eq:source_injection}
\end{equation}
Here $\sigma(t)$ is the time-dependent factor sampled from noise schedule. Our ablation studies confirm that this factor is critical; removing it causes the model to ignore the detail source structure, while replacing addition with channel concatenation degrades editability (see \cref{tab:cond}).

\textit{Reference Image Control (Sequence Concatenation):} 
Conversely, to enforce high fidelity to a reference object, we treat the reference image as an extension of the visual context. The reference image $\mathbf{x}_{ref}$ is patch-embedded and \textit{concatenated} to the input sequence of the DiT. This effectively extends the spatial-temporal attention window, allowing the model to ``copy'' texture details from the reference directly.

\noindent\textbf{Training Objective.}
We employ Flow Matching~\cite{lipman2022flow} as the training objective to minimize the mean squared error between the predicted velocity field $\mathbf{v}_\theta$ and the ground-truth drift:
\begin{equation}
    \mathcal{L}_{\text{flow}} = \mathbb{E}_{t, \mathbf{z}_0, \mathbf{z}_1, \mathbf{c}} \left[ || \mathbf{v}_\theta(\mathbf{z}_t, t, \mathbf{c}) - (\mathbf{z}_1 - \mathbf{z}_0) ||^2 \right]
\end{equation}
where $\mathbf{z}_1$ is the target video latent, $\mathbf{z}_0$ is standard Gaussian noise, and $\mathbf{c}$ represents the multimodal conditioning signals.

\subsection{Training Curriculum}
\label{sec:training}
Training a billion-parameter video generation model with multimodal conditions is computationally intensive and prone to optimization difficulties. To ensure stable convergence and effective alignment, we adopt a multi-stage progressive training curriculum as follows:

\noindent\textbf{Stage 1. MLLM-DiT Alignment.}
In the initial phase, we freeze both the MLLM base weights and DiT backbone. We train only the bridge components: the LoRA adapters, the Query/Latent Connectors, and the learnable query tokens. This stage uses text-based editing triplets $(V_{src}, T_{inst}, V_{tgt})$ and focuses on establishing a semantic mapping, ensuring the Connectors can translate MLLM representations into a format interpretable by the DiT's cross-attention blocks.
In this stage, the training data consists exclusively of high-quality image editing tasks sourced from GPT-Image-Edit~\cite{gpt-image-edit} and NHR-Edit~\cite{nhredit}, which provide a efficient way to align semantic space~\cite{pan2025transfer}.

\noindent\textbf{Stage 2. Instructional Tuning.}
We subsequently unfreeze the DiT layers to enable joint optimization. The model continues training on text-based editing triplets $(V_{src}, T_{inst}, V_{tgt})$ from large-scale instructional image and video editing datasets.
The training data for this stage combines the image datasets from Stage 1 with our curated instuction video editing subset (filtered with EditScore $\ge$ 6) as detailed in~\cref{sec:data_pipeline}.
This stage is crucial for learning general editing primitives (e.g., object removal, style transfer). To improve efficiency, we employ a resolution curriculum, initializing training with low-resolution clips ($480$p) and progressively scaling to higher resolutions ($720$p).

\noindent\textbf{Stage 3. Reference-Guided Fine-tuning.}
In the final stage, we introduce our curated \datasetname~dataset to unlock precise visual control.
We train on a mixture of instruction editing data from Stage 2 and new reference-guided quadruplets $(V_{src}, T_{inst}, I_{ref}, V_{tgt})$. 
This stage refines the model's ability to utilize the reference tokens for fine-grained textural transfer, ensuring the generated content aligns with user-provided visual examples.
During all training stages, we set the maximum number of frames sampled from video to 81.

\section{Experiments}

\begin{table*}[!t]
\centering
\caption{Comparison on RefVIE-Bench with \textit{Gemini-3-Flash} as Judge Model.}
\label{tab:ref_bench}
\setlength{\tabcolsep}{4pt}
\renewcommand{\arraystretch}{1}
\resizebox{\textwidth}{!}{%
\begin{tabular}{lcccccccccc}
\toprule
\textbf{Model} &
\makecell{\textbf{\#Params.}\textbf{(DiT)}} &
\multicolumn{4}{c}{\textbf{Subject Reference}} &
\multicolumn{4}{c}{\textbf{Background Reference}} &
\makecell{\textbf{Overall}$\uparrow$} \\
\cline{3-6} \cline{7-10}
&
&
\makecell{\textbf{Identity}\\\textbf{Consist.}} &
\makecell{\textbf{Temporal}\\\textbf{Consist.}} &
\makecell{\textbf{Physical}\\\textbf{Consist.}} &
\makecell{\textbf{Subj.}\\\textbf{Avg.}} &
\makecell{\textbf{Refer.}\\\textbf{Sim.}} &
\makecell{\textbf{Matting}\\\textbf{Quality}} &
\makecell{\textbf{Video}\\\textbf{Quality}} &
\makecell{\textbf{BG}\\\textbf{Avg.}} &
 \\
\midrule
\rowcolor{oran_tab} \textcolor{black!65}{Runway Aleph }&
\textcolor{black!65}{-} &
\textcolor{black!65}{3.79} & \textcolor{black!65}{3.65} & \textcolor{black!65}{3.58} & \textcolor{black!65}{3.67} &
\textcolor{black!65}{3.33} & \textcolor{black!65}{2.81} & \textcolor{black!65}{2.58} & \textcolor{black!65}{2.91} &
\textcolor{black!65}{3.29} \\
\rowcolor{oran_tab} \textcolor{black!65}{Kling-O1~\cite{team2025kling}} &
\textcolor{black!65}{-} &
\textcolor{black!65}{4.75} & \textcolor{black!65}{4.66} & \textcolor{black!65}{4.60} & \textcolor{black!65}{4.67} &
\textcolor{black!65}{3.95} & \textcolor{black!65}{3.21} & \textcolor{black!65}{2.75} & \textcolor{black!65}{3.30} &
\textcolor{black!65}{3.99} \\
\rowcolor{gray_tab} UniVideo~\cite{wei2025univideo} &
14B &
\underline{4.17} & \textbf{3.79} & \textbf{3.59} & \textbf{3.85} &
3.13 & 2.55 & 2.28 & \underline{2.65} &
\textbf{3.44} \\
\rowcolor{gray_tab} ReCo-Ref~\cite{Reco} &
2.1B &
3.65 & 2.95 & 2.70 & 3.10 &
\underline{3.35} & \textbf{2.65} & \underline{2.38} & \textbf{2.79} &
3.00 \\
\hline
\rowcolor{blue_tab} \textbf{Ours} (Stage-3 Insturuct-Reference) &
5B &
3.51 & 2.96 & 2.91 & 3.13 &
\textbf{3.40} & \underline{2.58} & \textbf{2.40} & \textbf{2.79} &
2.96 \\
\rowcolor{blue_tab} \textbf{Ours} (Stage-3 Reference. only) &
5B &
\textbf{4.28} & \underline{3.61} & \underline{3.55} & \underline{3.81} &
3.33 & 2.35 & 2.08 & 2.58 &
\underline{3.40} \\
\bottomrule
\end{tabular}%
}
\end{table*}

\begin{figure*}
    \centering
    \includegraphics[width=0.95\linewidth]{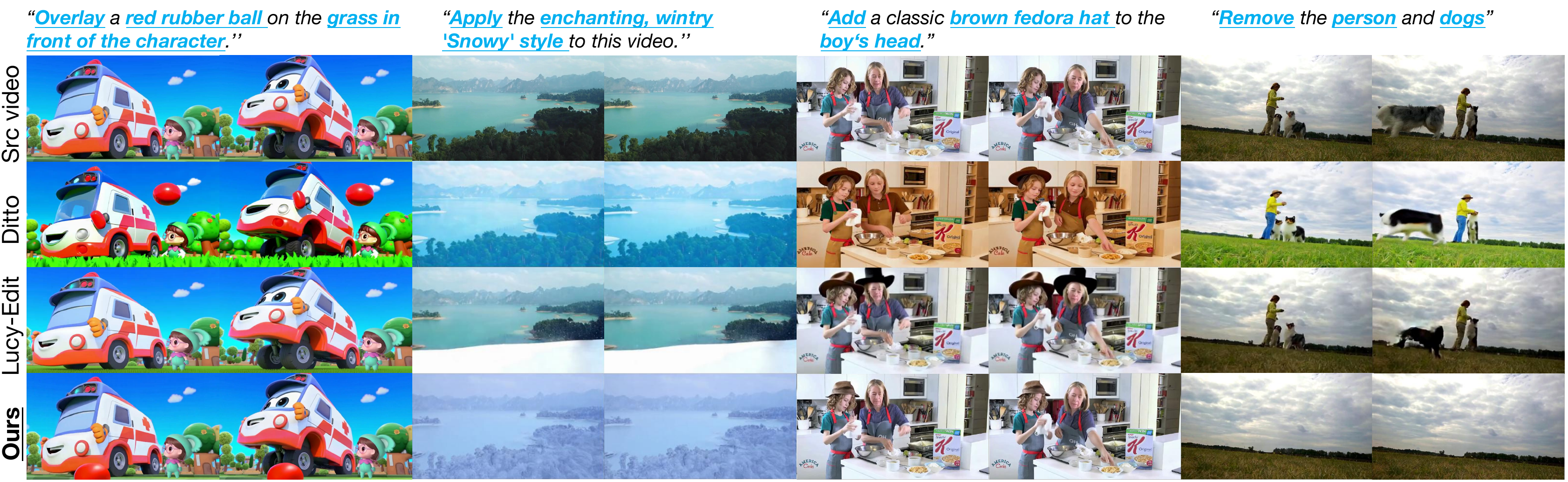}
    \caption{Qualitative results in OpenVE-Bench and VIE-Bench. Please zoom in for more details.}
    \label{fig:instruction_result}
    \vspace{-10pt}
\end{figure*}

\subsection{Implementation Details}
For image data, we utilize triplets (instruction, source image, edited image) dataset from NHR-Edit~\cite{nhredit} and GPT-Image-Edit~\cite{gpt-image-edit}.
We use the pretrained Wan-TI2V-5B as the DiT
backbone and finetune it on open source dataset and our propose reference data for 12K steps.
We set the learning rate to 2 × 10-5, with a global batch size of 128 with gradient accumulation.
In stage 2, we sample image and instruction video data at 1:1 ratio and first train on 360K pixels and then 960K pixels for 10K steps.
In stage 3, we sample image and instruction video data and video with reference image at 2:1:1 ratio for 10K steps.
More details are in supplementary materials.

\subsection{Main Results}
\noindent\textbf{Instruction Editing.} We compare our model with existing state-of-the-art open-source models on OpenVE-Benchmark~\cite{openve}, including VACE~\cite{jiang2025vace}, OmniVideo~\cite{tan2025omni},
InsViE~\cite{wu2025insvie}, ICVE~\cite{liao2025context}, Lucy-Edit~\cite{lucyedit}, and DITTO~\cite{ditto}, as well as the closed-source model Runway Aleph.
The evaluation setting follows the original setting report in the paper.
As shown in \cref{tab:dataset_comparison}
, our method outperforms all open-source baselines with an Overall score of \textbf{3.11}, significantly surpassing the previous best, OpenVE-Edit 2.50. Notably, our model excels in \textit{Background Change}, achieving a score of {2.67} which surpasses even the proprietary Runway Aleph 2.62. Additionally, increasing inference resolution to $1280 \times 704$ and applying training curriculum yields consistent performance gains across all metrics.

\noindent\textbf{Instruction and Reference Guided Editing.}
We compare our method against leading commercial video generation models, specifically Runway Aleph and Kling-O1~\cite{team2025kling}. As summarized in Table~\ref{tab:ref_bench}, our model trained on the curated \textit{\datasetname} achieves an overall score of \textbf{3.40}. 
Notably, our model demonstrates competitive performance in \textit{Identity Consistency} 4.28 and \textit{Reference Similarity} 3.40. While the proprietary Kling-O1 model achieves higher absolute scores, likely attributable to its significantly larger parameter count and closed-source training corpus, we setup a state-of-the-art baseline for open-source reference guided video editing.

\begin{figure*}[ht]
    \centering
    \includegraphics[width=0.98\linewidth]{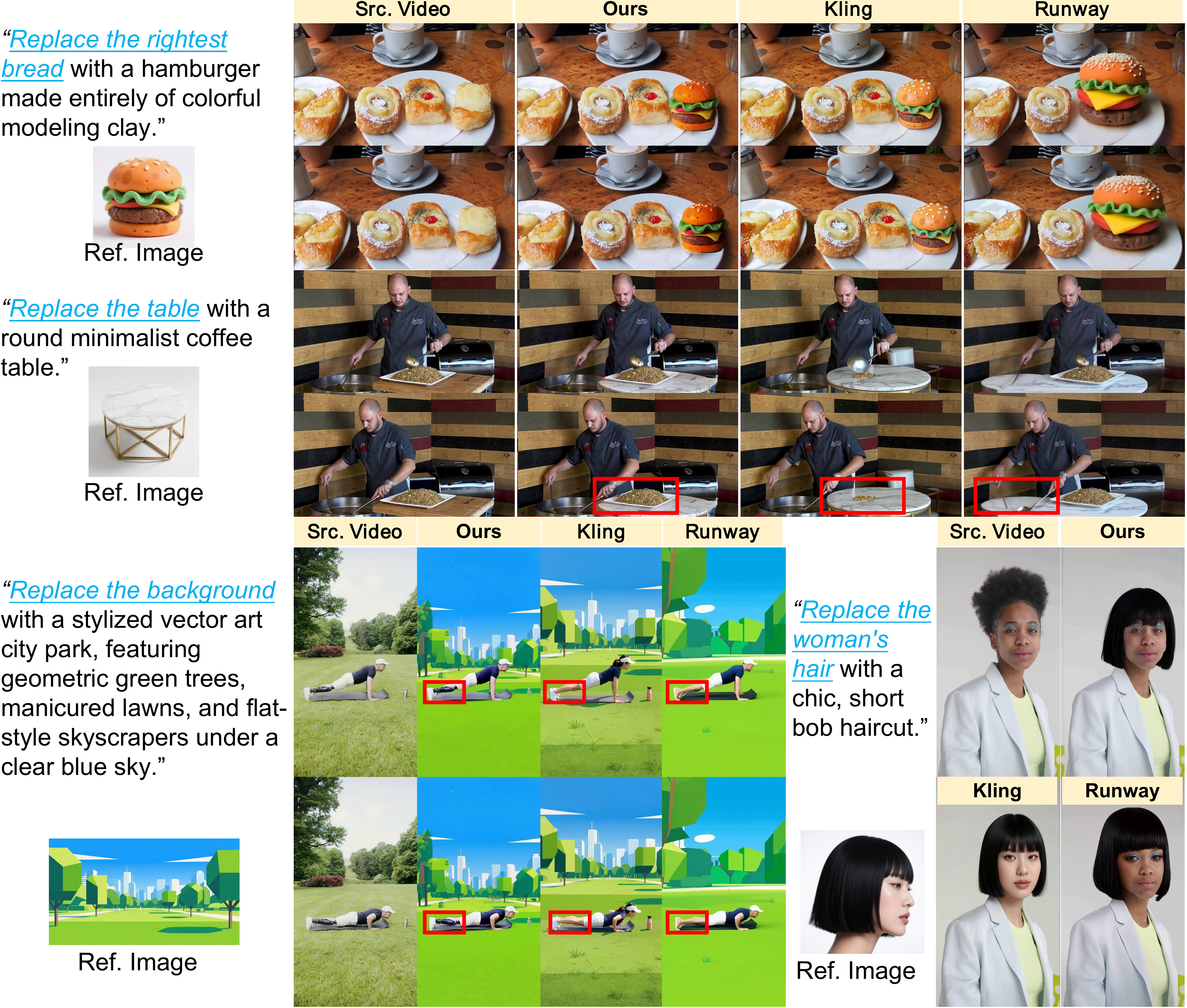}
    \vspace{-5pt}
    \caption{Qualitative results in our proposed RefVIE-Bench. Please zoom in for more details.}
    \vspace{-8pt}
    \label{fig:reference_result}
\end{figure*}

\subsection{Qualitative Results}
\cref{fig:instruction_result} and~\cref{fig:reference_result} present visual comparisons across diverse editing tasks. Our model exhibits superior instruction following and reference preservation.
\textit{Instruction Following:} Our model accurately captures visual semantics from the source and reference. For example, it correctly localizes the hat (\cref{fig:instruction_result}, Col. 3) and the table (\cref{fig:reference_result}, Row 2).
\textit{Reference Consistency:} As shown by the red bounding boxes in \cref{fig:reference_result} (Row 3), our method maintains high subject consistency even during drastic background style changes. 
\textit{More qualitative results are provided in the appendix and supplementary material.}

\subsection{Ablation Studies}
\noindent\textbf{Condition Design.} We conduct an ablation study on the model structure of source video input conditioning, comparing channel concatenation with feature addition strategies. As shown in \cref{tab:cond}, Channel Concat performs poorly, while sharing patch embeddings significantly degrades results to 1.01, confirming the need for independent feature extraction. The `Add w/ time-dependent factor' configuration proves most effective, outperforming the baseline in instruction tasks including remove (2.63) and add (2.01).

\begin{table}
\centering
\caption{Ablation on Conditional Design and using Image data.}
\resizebox{\linewidth}{!}{%
\begin{tabular}{lcc}
\hline
\textbf{Method} & \textbf{Score@Remove} $\uparrow$ & \textbf{Score@Style} $\uparrow$ \\ \hline
Add w/ time-dependent factor. & \textbf{2.63} & \textbf{4.07}\\
Add wo/ time-dependent factor. & 2.58 & {4.05}\\
Add (Shared Patch Embedding) & 1.01 & {1.00}\\ 
Channel Concat & 2.08 & 3.82\\
\hline
\midrule
Baseline (default) & \textbf{2.84}  & 3.98\\
w/o Alignment & 1.47 & 3.01 \\
w/o Image Co-train & 2.58 & \textbf{4.07}\\
\hline
\end{tabular}
\label{tab:cond}
}
\vspace{-5pt}
\end{table}

\noindent\textbf{Training Curriculum.} We perform an ablation study to validate the necessity of our progressive training stages, with results summarized in \cref{tab:cond}. 
First, skipping the \textit{Alignment} stage leads to a catastrophic performance drop, confirming that establishing a coarse semantic mapping between the MLLM and DiT is a prerequisite for effective instruction following. 
Second, excluding \textit{Image Co-training} degrades performance on structural tasks (Removal score 2.84 $\to$ 2.58). This indicates that while video-only training can achieve high style scores 4.07, it lacks the fine-grained spatial supervision provided by image editing datasets, which is essential for complex local manipulations.

\noindent\textbf{Reference Condition Design.} 
To validate the effectiveness of our dual-connector design, we analyze the impact of different condition injection pathways in the MLLM.
As shown in Table~\ref{tab:ab_query}, relying solely on learnable instructional queries yields a baseline score of 3.20. While queries effectively capture the high-level editing intent, they often struggle to preserve fine-grained visual details.
By incorporating the Reference Latent features via the Latent Connector, we explicitly inject dense semantic priors from the reference image into the context. This addition improves the score to 3.30, demonstrating that combining sparse instructional queries with dense visual latents is essential for achieving high-fidelity reference adherence.
\begin{table}
\centering
\caption{Ablation on architecture choice.}
\resizebox{0.7\linewidth}{!}{%
\begin{tabular}{ccc}
\hline
{Query } & {Ref. Latent } & \textbf{Score@Subject} $\uparrow$ \\ \hline
\ding{51} & \ding{55} & 3.20\\
\ding{51} & \ding{51} & 3.30\\
\hline
\end{tabular}
}
\label{tab:ab_query}
\vspace{-5pt}
\end{table}

\section{Conclusion}

This work addresses the critical scarcity of high-quality data for reference-guided video editing. We introduce a scalable pipeline to synthesize {\datasetname}, transforming existing video pairs into rich instruction-reference quadruplets, and establish RefVIE-Bench to standardize evaluation in this domain. Leveraging this data, our unified edit model \textbf{\modelname} achieves state-of-the-art performance, effectively bridging the gap between user intent and video generation. We believe this data-centric approach lays a solid foundation for more controllable and accessible video content creation.

\bibliography{main}
\bibliographystyle{icml2026}

\newpage
\appendix
\onecolumn
\section{Outlines}
The supplementary material presents the following sections to strengthen the main manuscript:
\begin{itemize}
    \item \cref{sec:append_dataset} details the Data Construction Process with Workflows.
    \item \cref{sec:append_bench} details the Benchmark results.
    \item \cref{sec:append_vis} provides additional Visualizations of the Reference dataset.
    \item \cref{sec:append_comp} shows further Quantitative and Qualitative Comparisons with SoTA methods.
    
\end{itemize} 

\section{Dataset Details}
\label{sec:append_dataset}

The following prompt is used for MLLM grounding and Reference image socre filtering.

\prompt{Editing Area Grounding}{
Given the image edit prompt and the source image and the edited image, generate the bounding box of the edited area in the edited image.
Output format:
\{
    "image\_2": [\{"bbox": [x1, y1, x2, y2], "label": "object1"\},...],
\}

Edit prompt is: 
}{prompt:eg}

\prompt{Foreground Grounding}{
Generate the bounding box of the same foreground object or person between the first and second images.
Output format:
\{
    "image\_1": [\{"bbox": [x1, y1, x2, y2], "label": "object1"\},...],
    "image\_2": [\{"bbox": [x1, y1, x2, y2], "label": "object1"\},...],
\}
}{prompt:fg}

\prompt{VLM Reference Image Score}{
You are an expert image editing evaluator.
Given:

- an input edit instruction (add or replace),

- an input image,

- an edited image,

- a reference image representing the edited region,

evaluate whether the reference image is qualified.

\textbf{Evaluation Criteria}

1. Instruction-Follow Consistency

   - The reference image must accurately represent the result of the input edit instruction as shown in the edited image.
   
   - Object identity, attributes (color, shape, material, style), and edit type must be consistent.
   
   - No contradictions with the edited result.

2. Image Quality \& Focus

   - Clear, realistic, and artifact-free.
   
   - Coherent structure, plausible lighting and texture.
   
   - The main subject must be clear and not overwhelmed by distracting elements.

\textbf{Scoring}

Output one overall score (1-10).

\textbf{Final Output (JSON Only)}

\{"score": 1-10 integer\}

}{prompt:score2}

\section{Benchmark Details}
\label{sec:append_bench}

The following prompt is used for RefVIE-Bench evaluation using Gemini~\cite{comanici2025gemini}.

\prompt{Evaluation Prompt of Subject Reference Editing Tasks}{
You are a data rater specializing in reference-guided object manipulation in videos. You will be given a Reference Image (the object to insert/swap), an Original Video, and the Edited Video. Your task is to evaluate the editing effect on a 5-point scale from three perspectives, specifically checking if the new object in the video matches the identity of the reference image.

\textbf{Identity Consistency \& Compliance}

1. Object not swapped/added, or a completely unrelated object appears.

2. Object is changed, but looks nothing like the reference image (wrong color, shape, or class).

3. Object class is correct, but identity details (texture, specific markings, logos) differ significantly from the reference image.

4. High resemblance to the reference image; correct geometry and texture, with only minor variations in fine details.

5. Perfect identity transfer: The object in the video is indistinguishable from the reference image in terms of texture, structure, and style, while maintaining the correct pose for the scene.

\textbf{Temporal Consistency \& Texture Fidelity}

1. The new object deforms, melts, or changes shape uncontrollably across frames.

2. Texture "swims" or flickers; resolution drops significantly compared to the rest of the video; object vanishes in some frames.

3. Object is stable in form, but texture details blur or shift slightly during motion; style looks somewhat pasted-on.

4. Object is structurally solid and texture is consistent; minor edge shimmer or noise visible only on close inspection.

5. Completely temporally coherent; the object maintains rigid structure (or appropriate flexibility) and consistent texture details in every single frame, exactly like a real object.

\textbf{Physical Integration \& Tracking}

1. Object slides around (bad motion tracking); does not follow camera or scene movement; looks like a sticker on the screen.

2. Missing interactions: No shadows, reflections, or occlusion handling (e.g., object appears on top of things that should be in front of it).

3. Motion tracking is decent with slight drift; lighting is flat or generic; occlusion is roughly correct but imprecise.

4. Accurate tracking; lighting and shadows match the scene's direction and intensity; correct occlusion handling.

5. Physically flawless: Motion tracking, perspective changes, motion blur, shadows, reflections, and lighting interactions are indistinguishable from reality; the object feels physically present in the scene.

The second and third score should no higher than first score!!!

Example Response Format:

Brief reasoning: A short explanation of the score based on the criteria above, no more than 20 words.

Identity Consistency \& Compliance: A number from 1 to 5.

Temporal Consistency \& Texture Fidelity: A number from 1 to 5.

Physical Integration \& Tracking: A number from 1 to 5.

editing instruction is : {prompt}
Below are the reference image, original video, and edited video:
}{prompt:score}

\prompt{Evaluation Prompt of Background Reference Editing Tasks}{
You are a data rater specializing in video background replacement grading. You will be given a Reference Image, an Original Video (foreground subject), and the Edited Video (result). Your task is to evaluate the background replacement effect on a 5-point scale from three perspectives, paying close attention to the preservation of the foreground subject and the fidelity to the reference image.

\textbf{Reference Fidelity \& Preservation}

1. Background not changed, or the foreground subject is severely damaged/removed.

2. Background changed but bears no resemblance to the reference image; foreground edges are significantly cut off or distorted.

3. Background resembles the reference but lacks key details; foreground is mostly preserved but has noticeable missing parts or artifacts.

4. Background clearly matches the reference image structure and style; foreground subject is fully preserved with only minor edge errors.

5. Perfect execution: The background is an exact semantic and stylistic match to the reference image, and the foreground subject is preserved pixel-perfectly throughout the entire duration.

\textbf{Matting Quality \& Temporal Stability}

1. Severe flickering; the background or foreground jitters erratically; distinct "boiling" artifacts on edges.

2. Obvious seams, halos, or "green screen" outlines around the subject; background moves unnaturally or freezes while the camera moves.

3. Edges are generally stable but soft/fuzzy; minor flickering in complex areas (e.g., hair, transparent objects); background stability is acceptable.

4. Clean edges with minimal temporal noise; background motion aligns well with camera movement; casual viewers notice no matting errors.

5. Completely seamless composition; hair/transparency details are perfectly matted; background and foreground interact with perfect temporal stability in every frame.

\textbf{Visual Harmony \& Perspective}

1. Background looks like a flat 2D image pasted behind a 3D subject; severe perspective or lighting mismatch (e.g., shadows point wrong way).

2. Lighting clashes (e.g., sunny background, dark foreground); no depth integration; subject looks "floating."

3. Perspective and scale are roughly correct; lighting is neutral but doesn't explicitly match the new environment’s ambience.

4. Good environmental integration; foreground lighting tones reflect the new background; cast shadows are present and mostly accurate.

5. Photorealistic integration: Depth of field, motion blur, lighting, and color grading of the foreground perfectly match the reference background; the composite looks like a single, raw video capture.

The second and third score should no higher than first score!!!

Example Response Format:

Brief reasoning: A short explanation of the score based on the criteria above, no more than 20 words.

Reference Fidelity \& Preservation: A number from 1 to 5.

Matting Quality \& Temporal Stability: A number from 1 to 5.

Visual Harmony \& Perspective: A number from 1 to 5.

editing instruction is : {prompt}
Below are the reference image, original video, and edited video:
}{prompt:score3}

\section{Sample Visualization}
\label{sec:append_vis}
\begin{figure*}
    \centering
    \includegraphics[width=0.9\linewidth]{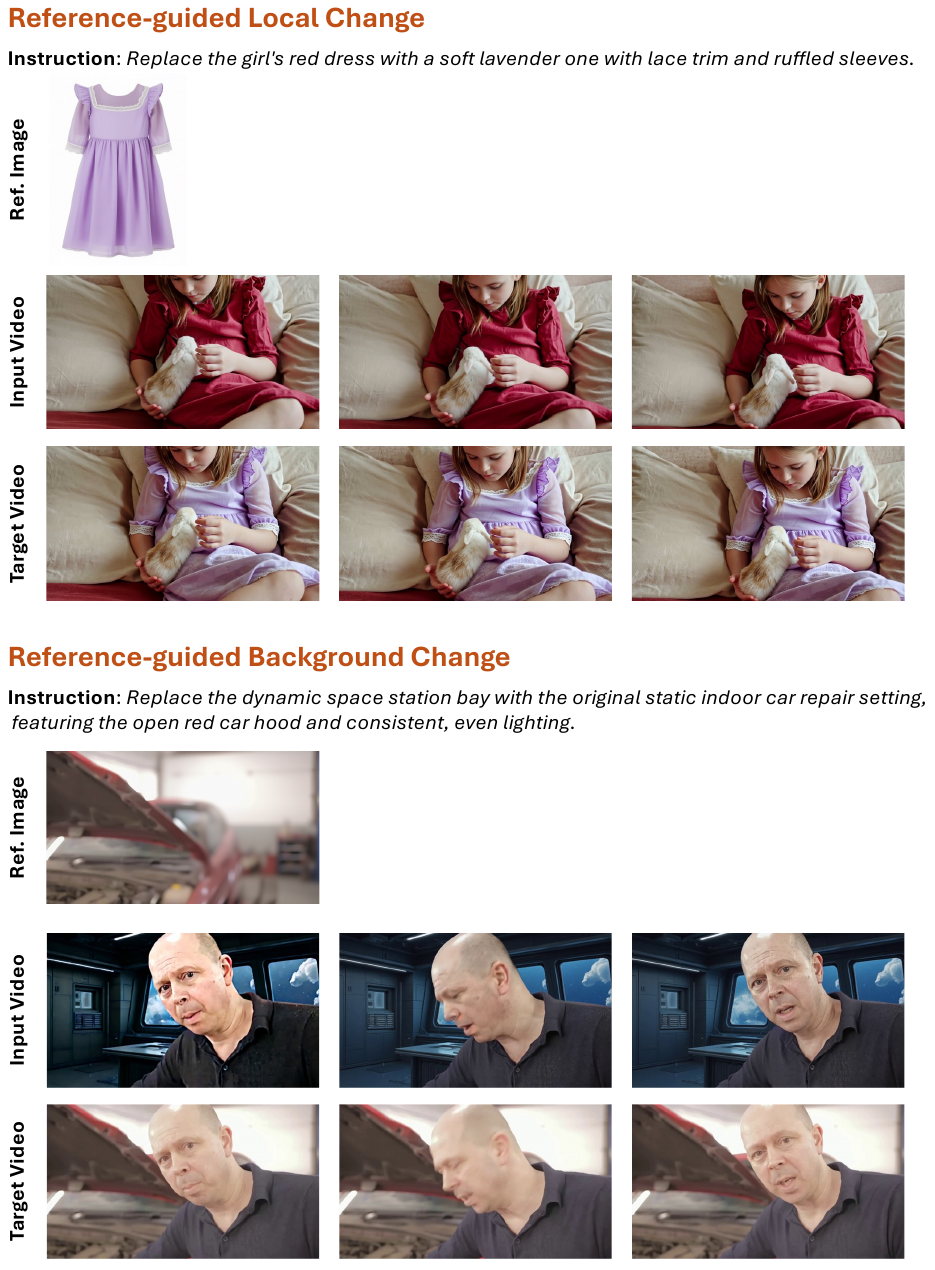}
\caption{Examples of our \datasetname.}
    \label{fig:supp_dataset_1}
\end{figure*}

\begin{figure*}
    \centering
    \includegraphics[width=0.9\linewidth]{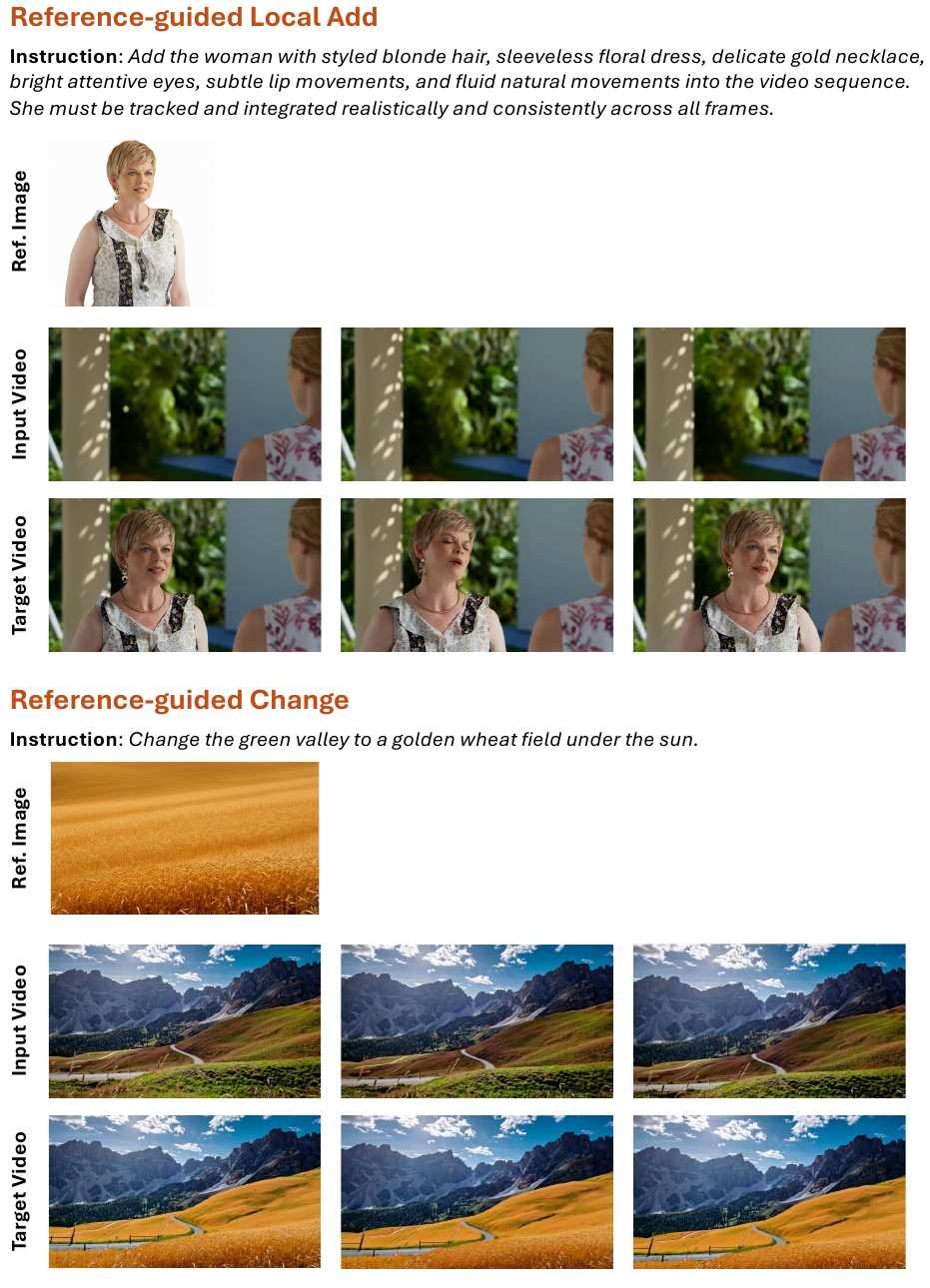}
\caption{Examples of our \datasetname.}
    \label{fig:supp_dataset_2}
\end{figure*}

\begin{figure*}
    \centering
    \includegraphics[width=0.84\linewidth]{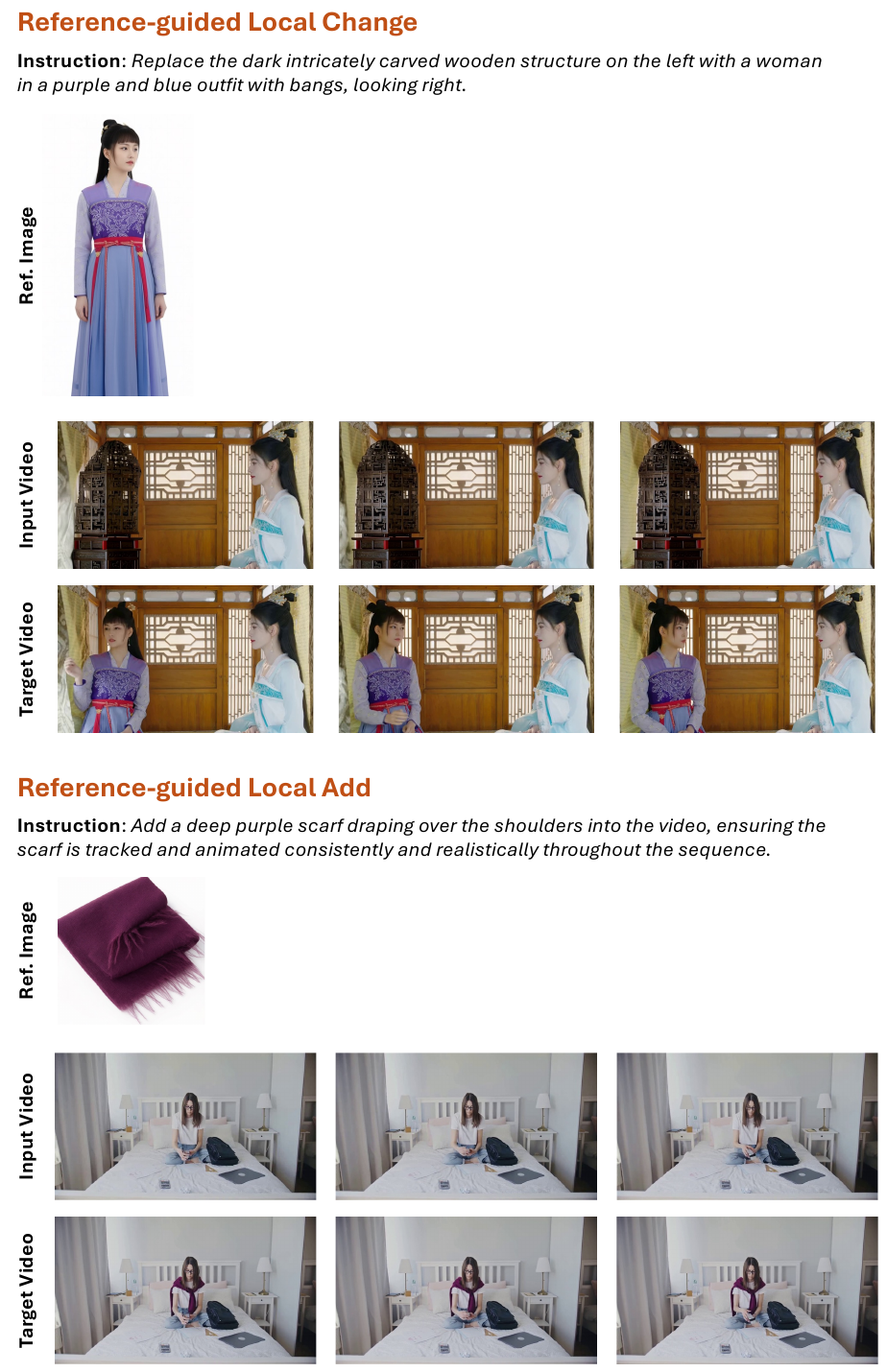}
\caption{Examples of our \datasetname.}
    \label{fig:supp_dataset_3}
\end{figure*}

In this section, we present additional samples from our high-quality \datasetname, as illustrated in Fig.~\ref{fig:supp_dataset_1}, Fig.~\ref{fig:supp_dataset_2}, and Fig.~\ref{fig:supp_dataset_3}.

\section{Qualitative Comparison}
\label{sec:append_comp}
\begin{figure*}
    \centering
    \includegraphics[width=0.9\linewidth]{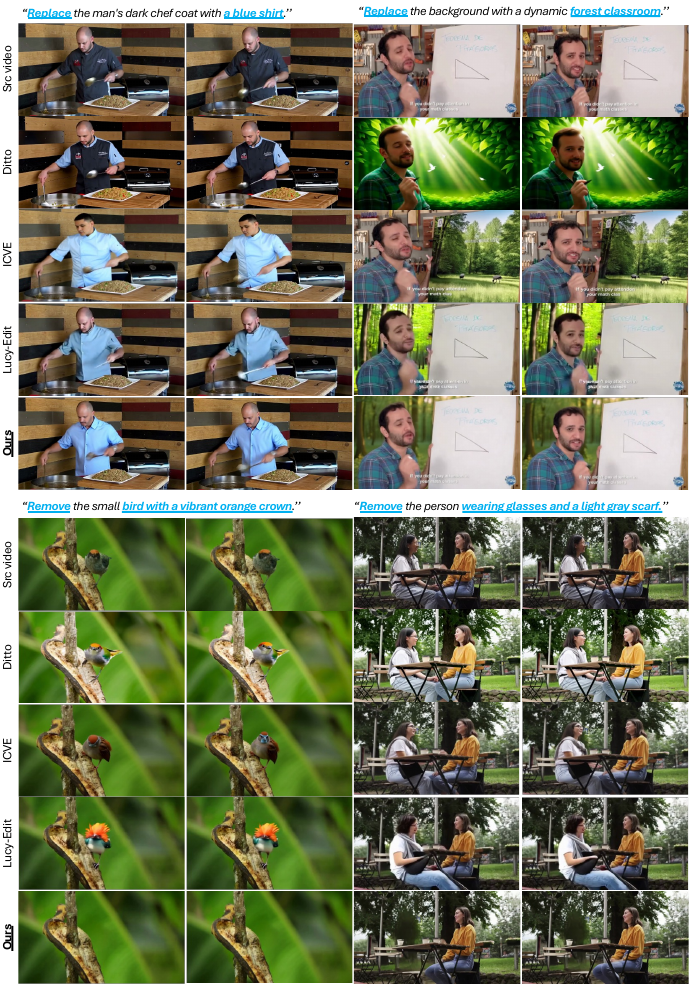}
\caption{Visual comparison on OpenVE-Bench.}
    \label{fig:supp_fig_2}
\end{figure*}

This section compares our method with SoTA methods using instruction-only input (Figs.~\ref{fig:supp_fig_2}--\ref{fig:supp_fig_3}). 
\textit{Instruction-reference-guided demo videos are provided in the supplementary material.}

\begin{figure*}
    \centering
    \includegraphics[width=0.9\linewidth]{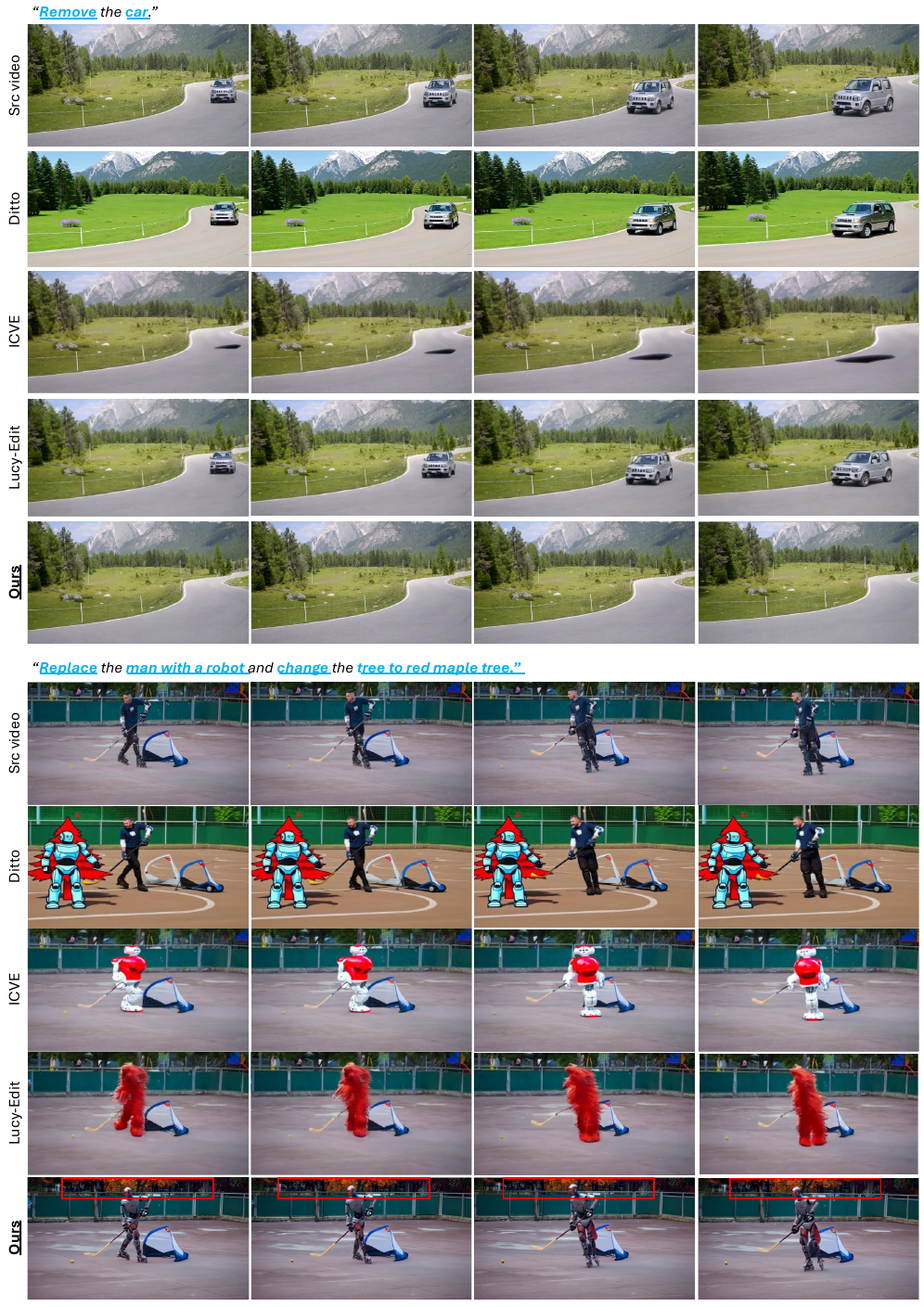}
\caption{Visual comparison on VIE-Bench. The bottom instruction not only replaces the man with a robot but also changes the tree in the background to a red maple tree. Only our method precisely follows this instruction, as highlighted in the \textcolor{red}{red} box.}
    \label{fig:supp_fig_3}
\end{figure*}


\end{document}